\RequirePackage{fix-cm}
\documentclass[smallcondensed]{svjour3}  
\smartqed 
\usepackage{graphicx}
\usepackage{amsmath}
\usepackage{amssymb}
\usepackage[linesnumbered,algoruled,boxed,lined]{algorithm2e}
\begin{document}
    
\title{Conformal Prediction based Spectral Clustering}

\author{Lalith Srikanth Chintalapati         \and
            Raghunatha Sarma Rachakonda
}
\institute{Lalith Srikanth Chintalapati,  Raghunatha Sarma Rachakonda \at
Department of Mathematics and Computer Science,\\ Sri Sathya Sai Institute of Higher Learning,\\ Vidyagiri, Prashanti Nilayam, Andhra Pradesh, India. \\
\email{lalithsrikanthc@sssihl.edu.in} 
}

\date{Received: date / Accepted: date}
\maketitle

\begin{abstract}
Spectral Clustering(SC) is a prominent data clustering technique of recent times which has attracted much attention from researchers. It is a highly data-driven method and makes no strict assumptions on the structure of the data to be clustered. One of the central pieces of spectral clustering is the construction of an affinity matrix based on a similarity measure between data points. The way the similarity measure is defined between data points has a direct impact on the performance of the SC technique.  Several attempts have been made in the direction of strengthening the pairwise similarity measure to enhance the spectral clustering. In this work, we have defined a novel affinity measure by employing the concept of non-conformity used in Conformal Prediction(CP) framework. The non-conformity based affinity captures the relationship between neighborhoods of data points and has the power to generalize the notion of contextual similarity. We have shown that this formulation of affinity measure gives good results and compares well with the state of the art methods.
\end{abstract}    
\section{Introduction and Literature Review}
\label{sect:intro} 
Clustering is an important unsupervised tool in understanding complex and unlabelled data.
It reveals the natural groups that make up the data, which in turn help the high-level data analysis operations.
Several techniques have been proposed in the field of data clustering which differ in their approach and design \cite{jain2005data,ng2002spectral,shi2000normalized,xu2005survey}. After the surge of hierarchical and density-based clustering techniques during the '90s, Spectral Clustering (SC) methods have assumed prominence due to their ease of applicability, implementation, and very good theoretical foundation.

SC has become popular due to many advantages it has, such as the ability to cluster the non-convex data, requiring only the pairwise similarity,
without imposing any condition on the global structure of the data.
The idea in SC is to use the eigenvectors of the Laplacian matrix to find the representation of the data points in an appropriate dimensional space so that a naive K-means can cluster them easily.

An essential concept in the field of pattern recognition is to compare two data vectors. If this comparison can be performed accurately, high-level operations such as classification, regression, and clustering become easy to perform.
There are many distances in the literature to compare two data vectors.
Typically in SC, the Gaussian kernel is used to find the pairwise-similarity \cite{ng2002spectral}.
Some authors have proposed similarity are based on the density of the respective points \cite{zhang2011local}. In recent times, the similarity is being based on not just distance or density but also the additional neighborhood properties such as structural similarity \cite{yang2011spectral}. Expanding on this idea, we have proposed a novel similarity metric based on the neighborhood sets of the respective points. 
We propose to capture the relationship between points and also between their corresponding neighborhoods in order to find pairwise similarity.

 In order to achieve this, the Conformal Prediction framework proposed by Vovk et. al.\cite{vovk2014basic} has been used in this work. In SC-based methods, the data is mapped to a graph where the data points are represented as nodes and the pairwise similarity among them is modeled as edge-weights. The SC algorithm then uses the graph-cut strategy for partitioning the data into k groups.
The most commonly used definition of similarity in SC is given by Gaussian kernel as follows:
The similarity between two points $z_i$ and $z_j$ is defined as :
\begin{equation}
A_{ij } = exp\bigg(\frac{-|| z_i- z_j ||^2}{2\times \sigma^{2}}\bigg)
\label{sim}
\end{equation}
Using this as the basis, a wide variety of measures have been proposed by estimating the sigma parameter \cite{gu2009improved,zelnik2004self} $\&$ kernel-based improvement \cite{arias2017spectral,challapower,diao2015spectral}. Several techniques have also been proposed wherein the neighbourhood of a point is utilized to arrive at an optimal affinity matrix \cite{li2012constructing,natalianipowered,ye2016robust}. Since the neighbourhood captures the local characteristics of the data point, it strengthens the notion of similarity between points and thus gives better results in SC framework.

Some of the methods in the literature which improve the similarity metric by optimally estimating the $\sigma$ parameter have been discussed below.
In the case of clustering of data, which is a combination of multiple distributions, a single value of $\sigma$ cannot give us the best clustering.
Addressing the estimation of local $\sigma$ using local properties, Zelnik and Perona proposed Self-Tuning Spectral Clustering \cite{zelnik2004self}. They estimate the local sigma($\sigma_i$) at the point $z_i$ as the distance from the point to its $k^{th}$ nearest neighbour.
Extending this idea to use the local information, Gu and Wang \cite{gu2009improved} defined local scale $\sigma_i$ as the average of distances from the point to its k nearest neighbors.
Using this local sigma, an affinity is defined as
\begin{equation}
    A_{il} = \exp    \bigg( \frac{-d_{il}^2}{\sigma_i\sigma_l}\bigg)
\end{equation}
where $d_{il}$ is the distance between $p_i$ and $p_l \in N_i$.

Apart from scale estimation, some authors have proposed to improve the affinity definition in SC by incorporating density information into it.
Since varying densities in the data is a challenging problem in clustering, pairwise affinity which use the local densities have been proposed in the literature.
Yang et al.\cite{yang2011spectral} have proposed a density sensitive SC method, in which the affinity between two points is said to be high if there is a path between them, which lies in the high-density region.

Using the K-means based density estimator with the sub-bagging procedure, Beauchemin \cite{beauchemin2015density} has proposed a novel affinity metric. The asymptotic properties of K-means were used for density estimation in Wong et al.\cite{wong1980asymptotic}. Using the work of Wong and sub-bagging procedure, an accurate density estimation was performed for effective formulation of similarity metric. 


The following are some of the methods in the literature which incorporate the local neighborhood information into the affinity definition by extracting additional characteristics from the data.
Using the Common Nearest Neighbors(CNN) between two points as an additional characteristic, Zhang \cite{zhang2011local} et al. have proposed an affinity measure. The similarity measure in their work is defined as
    \begin{equation}
        A_{kl}=
        \begin{cases}
            \exp\big(\frac{-d_{kl}^2}{2\times\sigma^2\times(CNN_{ij}+1)}\big)
            & k \neq l \\
            0                                              
            &    k=l
        \end{cases}
    \end{equation}
where $z_k,z_l  \in$ S, the data set. $\sigma$ is the scale parameter and $CNN_{kl}$ is the number of common nearest neighbors between $z_k,z_l$.
The CNN feature incorporates the notion of spatial nearness of the points into the similarity measure
    
Diao et al. \cite{diao2015spectral} proposed a metric which incorporates spatial structure based projection distance into the similarity.
The authors define a local projection neighbourhood (LPN) for any two data points $z_k, z_l$, The points within this neighbourhood are then projected on to the line connecting the two points.
The adjusted projection distance is defined over the projected points and used to define a novel affinity metric.
    
An enhancement to the similarity using the concept of `Neighbor propagation' has been proposed by Li and Guo \cite{li2012constructing}.
The authors follow the following steps.
The distance matrix $D=(d_{ij})$ which incorporates all the pairwise distances is constructed. From D, using Gaussian similarity function \ref{sim} the affinity matrix $A =(a_{ij})$ is calculated.
Then the symmetric Neighbor relation matrix $B=(b_{ij})$ is constructed as follows:
\begin{equation}
    b_{ij}=1, \hspace{.2cm} b_{ji}=1\hspace{.2cm} \textit{if}\hspace{.2cm} d_{ij}<\epsilon\hspace{.2cm} \textit{else}\hspace{.2cm} 0
\end{equation}
where $\epsilon$ is a fixed threshold.
    
Using the neighbor propagation rule described below the matrices B and W are updated accordingly. The neighbor propagation rule : If $b_{ij}$=1, $b_{jk}$=1 and $b_{ik}$=0, then set $b_{ik}$=1 and $b_{ki}$=1, simultaneously, update $a_{ik}$ and $a_{ki}$ as $\min(a_{ij}, a_{jk})$.
    
Shared neighbors between the points have been used by Ye and Sakurai \cite{ye2016robust} to propose two novel similarity metrics.
The first metric utilizes the number of shared neighbors. Given two points $z_i$ and $z_j$ and $N_i$ and $N_j$ are their respective neighbourhoods, then the similarity is defined as follows:
\begin{equation}
    A_{ij}=\frac{|N_i \cap N_j|}{k}
\end{equation}
where $|N_i \cap N_j|$ is the number of shared nearest neighbors between $N_i$ and $N_j$ and k is the maximum number of nearest neighbors shared between points $z_i$ and $z_j$ in the directed KNN graph.
    
The second similarity they proposed is based on how close the shared nearest neighbors are with respect to each other.
The shared nearest neighbors in $N_i \cap N_j$ are weighed according to their orders relative to the data points $z_i$, $z_j$.
The orders of shared nearest neighbors with respect to the data points $z_i$, $z_j$ are used to weigh them and used to create a new similarity:
    
If $z_r\in N_i\cap N_j$ and $z_r$ is $l_r^{th}$ neighbor of $z_i$ and $j_r^{th}$ nearest neighbor of $z_j$, then the weight($w_{ij}$) of the shared nearest neighbors in $N_i \cap N_j$ is given as: 
\begin{equation}
    w_{ij}= \sum_{z_r \in N_i \cap N_j}(k-i_r+1)(k-j_r+1)
\end{equation}
The second pairwise similarity, SC-cSNN is defined as:
\begin{equation}
    A_{ij}=\frac{w_{ij}}{\max(w_{ij})}
\end{equation}
This work utilizes the characteristics of the shared nearest neighbors, such as the number of shared nearest neighbors and how close they are to each other, thereby capturing the ``closeness and local structure" of the data points.
    
Arais et al.\cite{arias2017spectral} enhanced the similarity of spectral clustering using the local PCA features.
The procedure for calculating the affinity is as follows:
For the given set of points S = \{ $z_1$, $z_2$, $\ldots$, $z_n$ \} $\in$ $R^l$ $\&$ and $n_0$ centers are widely selected, A covariance matrix $C_i$ from the neighborhood of each center $y_i$ is computed. The authors define $Q_i$, an orthogonal projection of $y_i$ onto space spanned by the top  `d' eigenvectors of the matrix $C_i$.
    
The authors define the affinity among the centers as:
\begin{equation}
    A_{ij} =    \exp\Big(\frac{−||y_i - y_j||^2}{\epsilon^2}\Big)\times
    \exp\Big(\frac{−||Q_i - Q_j||^2}{\eta^2}\Big)     
\end{equation}
The cluster centers are clustered and the remaining data points are assigned to the nearest cluster centers with respect to the Euclidean distance.

A robust gamma-powered similarity was proposed by Nataliani et al. \cite{natalianipowered}. They define similarity as follows:
\begin{equation}
    \label{sim_pg}
    A_{ij } = \exp    \bigg( \frac{-|| z_i- z_j ||^2}{\beta}\bigg)^\gamma \text{ for i}\neq\text{j and 0 otherwise}
    \end{equation}
where $\beta$ refers to the neighborhood size and $\gamma>0$ is the given power parameter. 
The parameter $\beta$ is calculated as:
$\beta=\max (\min_{ j \neq i } ||z_i - z_j|| ) $.
A method to find novel similarity using Power Ratio Cut (PRCut) was proposed by Challa et al.\cite{challapower}. A faster SC model which retains the same accuracy has been proposed. Initially, the data is preprocessed using the Minimum Spanning Tree (MST), thereby considerably reducing the size and time complexity of the algorithm.
Using the $\Gamma$-limit of the spectral clustering algorithms, authors arrive at a fast version of SC. A discretization scheme to calculate the $\Gamma$-limit of the algorithms is employed. The PRCut based algorithm is also able to scale with the increase in the size of the data sets.
From the summary of the methods described above, it can be observed that the local information plays a vital role in the construction of the affinity matrix.

Conformal Prediction (CP) is a method which determines precise levels of confidence in the predictions of a machine learning algorithm \cite{vovk2014basic}.
In this work, these concepts from the field of CP have been used to capture the local properties effectively. Using concepts such as Non-Conformity Measure (NCM) and P-Value (PV) in CP, a novel formulation of the similarity measure has been proposed.

The following is the brief discussion on the related works in conformal prediction. 
Eklund et al.\cite{eklund2015application} have proposed conformal prediction based method for drug discovery. QSAR modelling is a widely used technique to prioritize the compounds for experimental testing or to alert about the toxic properties of the compound.
An offline inductive conformal prediction framework was proposed and applied to data obtained from AstraZeneca. Due to the violation of the randomness assumption, the validity of the conformal predictor was weakened. A semi-offline conformal prediction was adopted in order to strengthen the validity of the conformal predictor. They have found that in comparison to the traditional QSAR procedure, conformal predictions are highly useful in the drug discovery process.

Laxhammar and Falkman \cite{laxhammar2011sequential} have proposed anomaly detection using conformal prediction. They proposed a parameter-light algorithm called Similarity-based Nearest Neighbor Conformal Anomaly Detector (SNN-CAD) for online learning and sequential anomaly detection in trajectory data. They also proposed two parameter-free dissimilarity measures based on Hausdorff distance. They have demonstrated that their algorithm has a low false alarm rate without performing parameter tuning.

Smith et al. \cite{Smith2014Anomaly} proposed a conformal prediction based solution for finding anomalous trajectories in the maritime domain. They have used kernel density estimation based non-conformity measure in their algorithm. 

Cherubin et al. \cite{cherubin2015conformal} have proposed a clustering method based on conformal predictors as a multi-class unsupervised learning problem and applied it to the classification of bot-generated network traffic. The idea of cluster creation proposed by them: the prediction set proposed by the CP is interpreted as a set of possible objects which conform to the data set. This prediction set might consist of several parts that are interpreted as clusters. The significance level is used to regulate the depth of the clusters' hierarchy. Authors used the \textit{neighbouring rule} to create the clusters and used Purity criterion for evaluating the cluster's accuracy.

In CP, $P\_value$ is defined as the measure of conformity of a point $z_i$ with a set $S$.
The non-conformity between point and a set is defined in multiple ways depending on the application.
A relationship between a point and a set is defined using the concept of individual non-conformity.  If $P\_value(z_i, S)$ is high, then $z_i$ conforms to $S$ or in other words could be similar to the points in $S$ as a whole. 
The notion of conformity has been used to define the similarity measure among data points. Our Contribution is the new definition of affinity measure based on non-conformity and P-value functions.
\subsection{Our Contributions}
A new approach to defining the affinity measure between data points has been proposed, which incorporates local characteristics of their neighbourhoods.
    
The main advantage of our work is that the similarity is based on the comparison between not just points but also their neighbourhoods, which gives us a greater sense of contrast between points.
    
The outline of the paper is as follows:    
Section \ref{SpectralClusteringAlgorithm} presents the theory of spectral clustering, conformal prediction and the motivation behind our algorithm. In section \ref{Methodology}, the methodology of our algorithm: modelling of data and the modified SC algorithm with proposed affinity measure have been presented. Section \ref{results} discusses the results of our algorithm in comparison with state of the art techniques in SC. The conclusion and future work are presented in section \ref{conclusion}.        
\section{Background}
\label{SpectralClusteringAlgorithm}
\subsection{Spectral Clustering Approach}
\label{SpectralClusteringApproach}
Let the given data set be represented as $Z=\{z_1,...,z_n\}$ in $R^l$. The data points in Z are modelled as nodes of a graph, and an undirected $\varepsilon$-graph G = (V, E), where $V=\{v_1,...,v_n\}$ is the set of vertices and E is the set of edges is constructed as follows:

If the distance between any two data points $z_i$ and $z_j$ is less than a given parameter $\varepsilon>0$, then the corresponding nodes $v_i,v_j$ in the graph G are connected via an edge, denoted as $(v_i,v_j)$ i.e. E $= \{(v_i,v_j)|$ $ d_{z_iz_j} < \varepsilon \hspace{.2cm} \forall v_i, v_j \in V \}$, where $d_{z_iz_j}$ is the distance between the data points $z_i$ and $z_j$.
Let each edge $(v_i,v_j)$ be weighted by  $s_{ij}\geq0$ which denotes the similarity between the data points $z_i$ and $z_j$(such as Gaussian similarity in Eq.\ref{sim}). 
    
The fundamental idea in SC is to find an optimal low dimensional embedding of the given data such that the clustering of data becomes easy. In order to find this embedding, the Eigen spectrum of the graph Laplacian is employed.
 In their work, Ng et al.\cite{ng2002spectral} have used the top $k$ eigenvectors of graph Laplacian to find the embedding of the data.
 The Laplacian matrix is evaluated as follows:
    The affinity matrix $A$ is defined as $A_{ij}=(s_{ij})$ where $s_{ij}$ is the similarity between $z_i$ and $z_j$. The degree matrix $D$ is defined as a diagonal matrix with $D_{ii}=\sum_j s_{ij}$. A normalized Laplacian matrix $L$ is constructed from $A$ and $D$ as: $L=D^{-1/2}AD^{-1/2}$. The top $k$ eigenvectors obtained from Eigen Value Decomposition of Laplacian $L$ are used to find $k$-dimensional embedding of the given data. Algorithm \ref{Algo1} lists out the steps in spectral clustering procedure as given in Ng et al. \cite{ng2002spectral}.
    \begin{algorithm}[H]
        \KwIn{A set of points $Z=\{z_1,...,z_n\}$ in $R^l$, parameters $k$, $\sigma$}
        \KwOut{k clusters of the given data}
        \begin{enumerate}
            \item Form the affinity matrix $ A \in R^{n \times n}$ defined by $A_{ij}=\exp(-||z_i -z_j||/2{\sigma}^2)$
            \item Define D to be the diagonal matrix whose $(i,i)$- element is the sum of i-th row of A, and construct the matrix $L=D^{-1/2}AD^{-1/2}$.
            \item Find $e_1 ,e_2 , \ldots, e_k$, the k largest eigenvectors of $L$ (chosen to be orthogonal to each other in the case of repeated eigenvalues), and form the matrix $X = [e_1 \hspace{.3cm} e_2 \hspace{.3cm} \dots\hspace{.3cm} e_k] \in R^{n \times k}$ by considering the eigenvectors as columns of the matrix.
            \item Form the matrix $Y$ from $X$ by re-normalizing each of X's rows to have unit length (i.e. $ Y_{ij} = X_{ij}/ (\sum_{j}X_{ij}^2)^{1/2})$.
            \item Treating each row of Y as a point in $R^k$, cluster them into $k$ clusters via K-means or any other algorithm (that attempts to minimize distortion).
            \item Finally, assign the original point $s_i$ to cluster j if and only if row i of the matrix
            $Y$ was assigned to cluster j.        
        \end{enumerate}
        \caption{Simple spectral clustering}
        \label{Algo1}
    \end{algorithm}

\subsection{Conformal Prediction and Clustering}
\label{cp}
In this subsection we briefly present the theoretical background of the Conformal Prediction(CP) framework\cite{vovk2014basic}.
Typically, CP is used to assign confidence measure on the predictions made by a machine learning algorithm.
One of the main requirements of CP is that points in $Z$ are exchangeable, which means it is order-independent.
Given a set of points $ Z= \{ z_1, \ldots, z_{n-1} \}$, a new point $z_n$ and a significance level $\epsilon$. If $P$ is the prediction set with possible values which might be predicted for $Z$.
CP determines if $z_n$ comes from $P$ with an error on the long run of at most $\epsilon$.
In the CP framework, P-value is an important concept defined to encapsulate the relationship between a point $z_n$ and a set $Z$. The algorithm \ref{Algo2} details the steps in calculating the P-value. The point $z_n$ is said to conform to $Z$ if the P-value between them is above a threshold $\varepsilon$.

Given a new point $z_n$, non-conformity gives us how well does that conform with an existing set of points $Z$. The Non-conformity measure $NC$ is defined as:
\begin{equation}
    NC:    Z^{*} \times Z \longrightarrow R
\end{equation}
where $Z^{*}$ is the set of all possible subsets of Z.\\
There are many types of non-conformities proposed in the literature \cite{cherubin2015conformal}. 
The two popular non-conformity functions are based on K nearest neighbours (KNN) and Kernel Density Estimation (KDE). The KNN based non-conformity measure($NC$) is defined as:
\begin{equation}
    NC(z_{i}, Z)=\sum_{j=1}^{k}\delta_{ij}
\end{equation}
where $\delta_{ij}$ is the $j^{th}$ smallest distance between $z_i$ and the objects in $\{z_1, ... , z_n\}\setminus z_i$ and $k$ is the number of nearest neighbors.
        
The KDE based $NC$ is defined as:
\begin{center}
    $NC(z_
    i, Z) = -\Bigg(\frac{1}{nh^{d}}\sum_{j=1}^{n}K\Big(\frac{z_i-z_j}{h}\Big)\Bigg)$\\
\end{center}
where $K: R^d \longrightarrow R$ is the kernel function, $d$ is the number of features, $h$ is the kernel bandwidth and $K(u)$ is the Gaussian kernel:
\begin{equation}
    K(u) = \dfrac{1}{2\pi}e^{-\frac{1}{2}u^2}
\end{equation}
As mentio
The algorithm to calculate the P-value($P$) as given by Balasubramanian et al.\cite{balasubramanian2014conformal} is as follows:
\begin{algorithm}
\KwIn{$D_0 =\{z_1, ..., z_{n-1}\}$ is the given set of points, Non-conformity measure $NC$, Significance level $\varepsilon$ and the new point $z_i$}
\KwOut{P-value($P(z_i,D_0)$)}
\begin{enumerate}
    \item     Set provisionally $z_n=z_i$ and $D=D_0 \cup \{z_n\}$, $D =\{z_1, ..., z_n\}$
    \item     \textit for $ k \longleftarrow 1$ to n do\\
            $|$  $\alpha_k \longleftarrow NC(D\setminus z_k,z_k)$\\
            end
    \item $\tau=U(0,1);$
    \item \begin{equation}
                P(z_i, D_0)=\frac{\#\{i:\alpha_k>\alpha_n\} +\#\{i:\alpha_k=\alpha_n\}\tau}{n}
            \label{pn_equation}
            \end{equation}
\end{enumerate}
    \caption{Evaluation of P-value}
        \label{Algo2}
\end{algorithm}
The value $P$ gives us the measure of how similar or dissimilar the point is to a group of points.
Since the concept of non-conformity and P-value($P$) can be used to define the relationship between a point and a neighbourhood set of another point, we hypothesized that the value of $P$ would represent the similarity between two points. 
\subsection{Motivation} 
The methods of CP and SC were discussed in the previous sections. In Spectral clustering method, a metric is used for estimating the similarity between any two data points. In Conformal prediction framework, the conformity between a novel point and a set is calculated to evaluate the P-value. The conformity between the two points $x$ and $y$ can be further seen as the conformity between $x$ and neighbourhood of $y$ and vice versa. This relation between a point and a neighborhood set is a new way to formulate the similarity between two data points. Using the definition of $P$, we propose a definition of affinity in the next section.
\section{Methodology}
\label{Methodology}    
 In this section, we present the proposed affinity metric based on P-value.
\subsection{Construction of CPSC based affinity metric}
Given a set of data points, we model them as nodes of a graph G as described in section 
\ref{SpectralClusteringApproach}. 
The neighborhood set of the node $u$ is defined as: $N(u) = \{v | (u,v) \in E \}$. The parameter $\epsilon$ controls the sparseness of the neighborhood graph.
We propose two types of similarity using the non conformity measure and $P-value$. We define the similarity CPSC($A_{ij}$) between two points $z_i$, $z_j$ as the P-value between $z_i$ and the neighborhood of $z_j$. Using the definition of $P$ in Eqn. \ref{pn_equation} we define $A_{ij}$ as:
\begin{equation}
    A_{ij}=P(z_i,Nbd(z_j))
\end{equation}
From the definition of $A_{ij}$ we can see that it is not symmetric. The $A_{ij}$ need not be equal to $A_{ij}$. In order to propose a symmetric affinity metric, we propose a mean $P$ by taking average of $P(z_i,Nbd(z_j))$ and $P(z_j,Nbd(z_i))$. We denote this affinity by CPSCA($\hat{A}_{ij}$). It is defined as :\\
\begin{equation}
    \hat{A}_{ij}=\frac{P(z_i,Nbd(z_j))+P(z_j,Nbd(z_i))}{2}
\label{aijhat}
\end{equation}
Using the definition of P-value in sec. \ref{cp}, the value of P signifies the conformity of a point $z_i$ to another set S. In this test, the set S is a set of neighborhood points of another point $z_j$. If $z_i$ conforms with S, then $z_i$ is similar to $z_j$. Hence the NULL hypothesis is proved.
\subsubsection{Proposed algorithm}
We propose a hybrid similarity using the Gaussian and steering features based similarity (Eq. \ref{aijhat}). This is given as :
\begin{equation}
\label{finalA}
A_{ij } =\hat{A}(z_i,z_j)+exp\bigg(\frac{-|| z_i- z_j ||^2}{2\times \sigma^{2}}\bigg)
\end{equation}
We have employed this similarity in the SC algorithm provided by Ng et al.\cite{ng2002spectral}. The following is the modified SC algorithm with the proposed affinity:\\
\begin{algorithm}[H]
\SetAlgoLined
\KwIn{Given data points, $k$}
\KwOut{The labels of the input data points}
    From the data points, construct an affinity matrix $A$ using similarity $A_{ij}$ in Eq. \ref{finalA}
    From $A$, a normalized Laplacian matrix $L$ is constructed.\\
    Top k eigenvectors of L (k is the number of clusters) are computed.\\
    The vectors obtained are further placed as columns to form matrix $\hat{P}$, and rows of $\hat{P}$ represent the original data points.\\
    Rows of the matrix $\hat{P}$ are clustered using the K-means algorithm.\\
    Original points are labelled based on results of the K-means clustering.            
\caption{Conformity Prediction based Spectral Clustering(CPSC)}
\end{algorithm}
    
\section{Experiments and Results}
\label{results}
In this section, we present the framework for experimentation, the results obtained and the analysis of the proposed algorithm. The experiments were conducted on three different types of data sets. 

We compared our technique with the following methods: spectral clustering algorithm (NJW) by Ng et al. \cite{ng2002spectral}(2002), Neighbor propagation (NP) (2012) method proposed by Li and Guo. \cite{li2012constructing}, Shared Nearest Neighbors (SNN) (2016) based method proposed by Ye and Sakurai. \cite{ye2016robust}, Powered Gaussian (PG) (2017) based spectral clustering by Nataliani et al. \cite{natalianipowered}, Spectral clustering using Local PCA (LPCA)\cite{arias2017spectral} (2017), Powered Ratio Cut (PRCUT)\cite{challapower} (2018). 
    
We have presented the results using three types of metrics. They are: Adjusted Rand Index (ARI)\cite{rand1971objective}, Normalized Mutual Information (NMI)\cite{strehl2003cluster} and Clustering Error (CE)\cite{jordan2004learning}. The definition and detailed explanation of these metrics are given in the Appendix.
As the clustering accuracy with respect to the ground truth increases, values of ARI and NMI metrics will go closer to 1, and the value of CE metric will become smaller (tending to 0).
\subsection{Parameter Estimation}
The proposed CPSC algorithm has two input parameters $\varepsilon$ and $k$, where $\varepsilon$ denotes neighbourhood size of a point, and $k$ is used in the non-conformity function. Estimation of  $\varepsilon$ is done as follows: \\
The given data is normalized, and distance(Euclidean) between all the data points is calculated. Let $max_{dist}$ represent the maximum of all the pairwise distances among the data points. We then run the proposed spectral clustering CPSC on the set of values [0, $max_{dist}$] with step size .01.
The input parameter $k$ is taken in the range of 1 to 30.
We then evaluate the output of these clustering runs with an internal metric Silhouette Index \cite{kaufman2009finding}. The combination of input parameters ($\varepsilon$,$k$) for which we obtain the higher value of SI value is considered as the output of the CPSC algorithm. 
Li et al.\cite{li2012constructing} proposed automatic estimation of $\varepsilon$ in their algorithms. In our experiments, we observed that the methods mentioned above do not provide an accurate estimate of $\varepsilon$. Hence we used an internal metric to arrive at optimal parameters, making it a completely unsupervised way of selecting parameters. The following subsection details our experiments on synthetic datasets.
\subsection{Synthetic data sets}
\begin{figure}
\centering
\includegraphics[width=\linewidth]{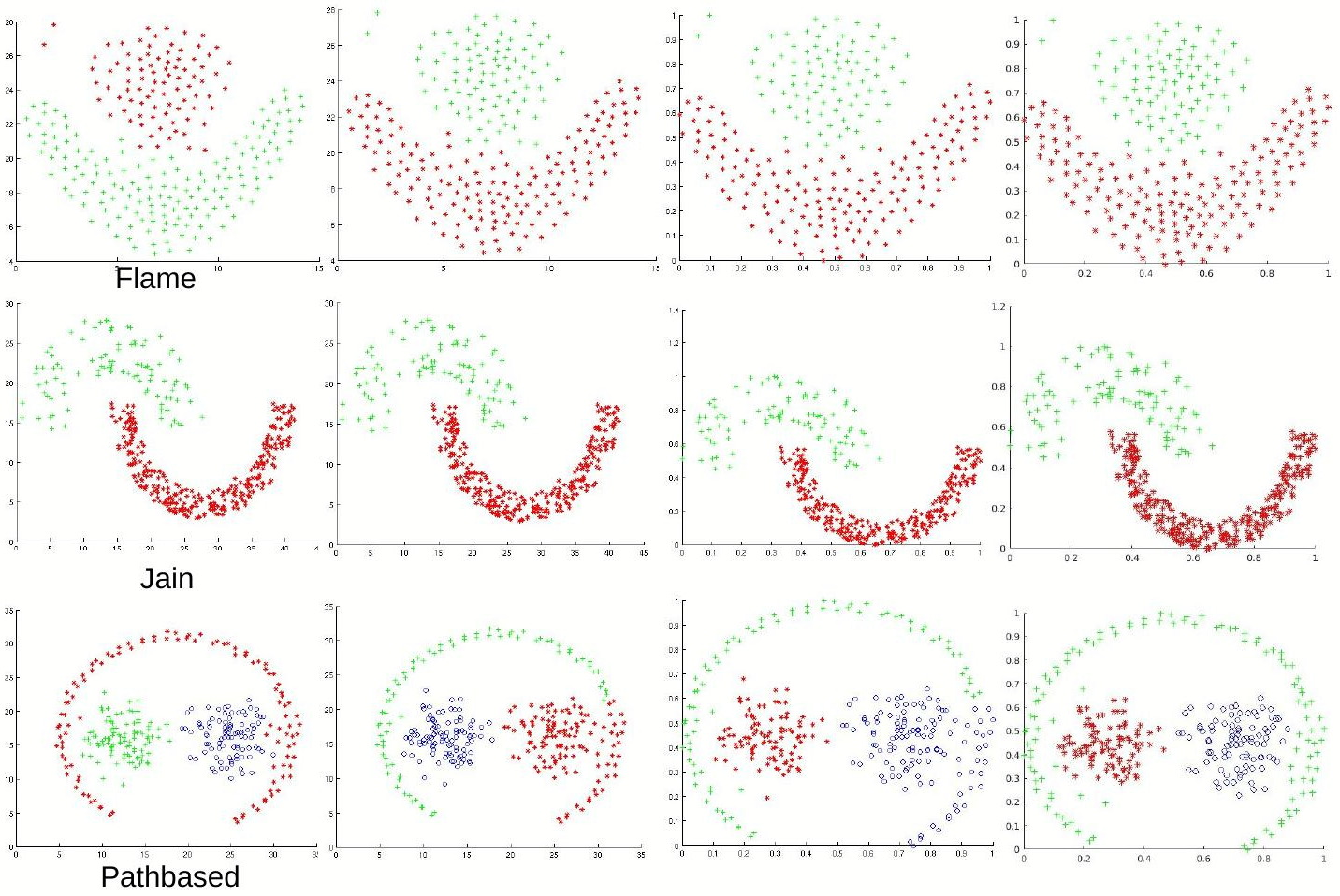}
\includegraphics[width=\linewidth]{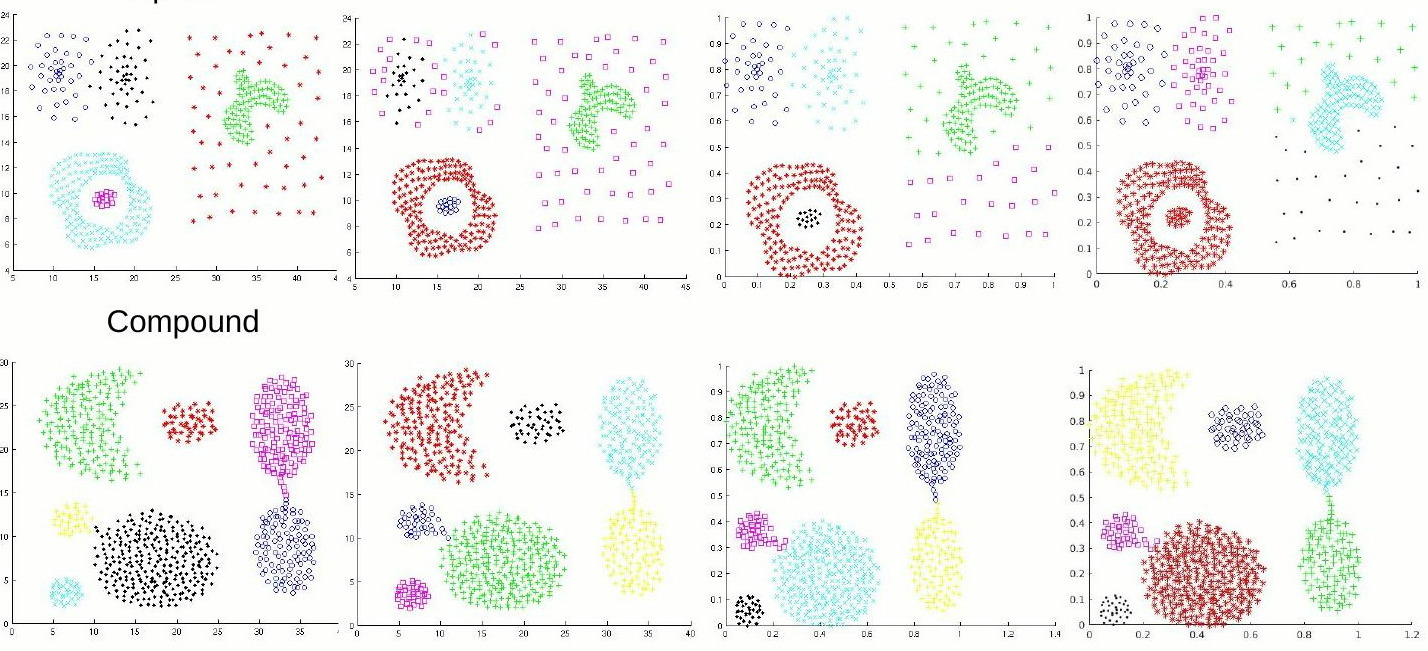}
\includegraphics[width=\linewidth]{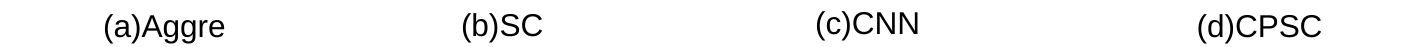}
\caption[Summation Index]{Comparison of results of various methods with CPSC. (a.) Original data sets (b.) Results of NJW (c.) Results of CNN (d.) Results of CPSC} 
\label{fig:syntheticResults}
\end{figure}

In the first category of data sets, synthetic data sets, we have considered four two dimensional synthetic data sets. They are Compound\cite{zahn1971graph}, Aggregation\cite{gionis2007clustering}, Flame\cite{fu2007flame} and Pathbased\cite{chang2008robust}.
Each data sets presents challenges such as varying local density, connectedness of data, etc. Results of CPSC method with other state of the art methods are displayed in Fig. \ref{fig:syntheticResults} and Tables \ref{table:1}, \ref{table:2}, \ref{table:3}.    
            
\begin{table}
\caption{Results of ARI metric on Synthetic Data sets}
\label{table:1}
\centering    \begin{tabular}{|p{1.6cm}|p{1.2cm}|p{1.2cm}|p{1.2cm}|p{1.2cm}|p{1.4cm}|}
\hline
& & & Methods & &\\
\hline        
\hline
Data sets    &    SC    &    NP    & SNN &    PG    & CPSCA\\
\hline 
Flame & 0.5116 & \textbf{0.9666} & 0.4093 & 0.8543 & 0.9501 \\
Compound & 0.5156 & 0.7992 & 0.5171 & 0.5155 & \textbf{0.8156}\\
Aggregation & 0.8372 & 0.8792 & 0.8418 & \textbf{0.9670} & 0.5318\\
Pathbased & 0.4797 & 0.5380 & 0.5176 & 0.4797 & \textbf{0.5686}\\
\hline
\end{tabular}             
\end{table}
\begin{table}
\caption{Results of NMI metric on Synthetic Data sets}
\label{table:2}
\centering
\begin{tabular}{|p{1.6cm}|p{1.2cm}|p{1.2cm}|p{1.2cm}|p{1.2cm}|p{1.5cm}|}
\hline
& & & Methods & &\\
\hline        
\hline
Data sets    &    NJW    &    NP    & SNN &    PG    & CPSCA\\        \hline 
Flame &\textbf{1.0000} & 0.9355 & 0.8991 & 0.9269 & 0.9269 \\
Compound & 0.9171 & 0.8573    & 0.7694 & 0.9119&0.8526\\
Aggregation & 0.9824 & 0.9342& 0.9799 & 0.9824 & \textbf{0.9851}\\
Pathbased &    0.7825    &    0.7664    & 0.6082 & 0.7664& \textbf{0.9125}\\
\hline
\end{tabular}             
\end{table}    
\begin{table}
\caption{Results of CE metric on Synthetic Data sets}
\label{table:3}
\centering
\begin{tabular}{|p{1.6cm}|p{1.2cm}|p{1.2cm}|p{1.2cm}|p{1.2cm}|p{1.5cm}|}
\hline
& & & Methods & &\\
\hline        
\hline
Data sets    &    NJW    &    NP    & SNN &    PG     & CPSCA\\        \hline 
Flame & \textbf{0.0000} & 0.0083 & 0.0125 & 0.0083 &  0.0083 \\
Compound & 0.0526 & \textbf{0.0326} & 0.3308 & 0.0702  & 0.1328\\
Aggregation & 0.0063 & 0.1320    & 0.0076 & 0.0063 & \textbf{0.0051}\\
Pathbased &    0.1133 & 0.1300 & 0.1933 & 0.1300 & \textbf{0.0233}\\
\hline
\end{tabular}             
\end{table}            
\subsection{Real data sets}
The second category of data we have considered is UCI real data\cite{Lichman:2013}. These data features are extracted from real scenarios and have varied number of features  such as number of clusters (2 to 8), number of data points (74 to 768). 
Characteristics of five data sets we considered are displayed in Table \ref{table:36}. 
Results of CPSC in comparison with other SC methods are given in Tables \ref{table:4}, \ref{table:5}, \ref{table:6}. The results show that in most of the cases, CPSC outperforms the other methods. 
\begin{table}[!h]                                     
\caption{Attributes of real UCI data sets}        
\label{table:36}
\centering
\begin{tabular}{ |p{3.2cm}| p{.8cm}| p{.8cm}| p{.8cm}| p{.8cm}| p{.8cm}|}    \hline
Dataset & Wine & Glass & Iris & Ion & Sonar\\        \hline 
No of instances & 178 & 214 & 150 & 351 &     208\\  
No of attributes & 13 & 9 & 4 & 34 & 60\\
No of clusters & 3 & 6 & 3 & 2 & 2\\%
\hline
\end{tabular}                                                    \end{table}                
\begin{table}
\caption{Results of ARI metric on UCI real Data sets}
\label{table:4}
\centering    \begin{tabular}{|p{1.2cm}|p{1.2cm}|p{1.2cm}|p{1.2cm}|p{1.2cm}|p{1.2cm}|p{1.2cm}|p{1.5cm}|}
\hline
& & & Methods & & & &\\
\hline        
\hline
Data sets & NJW & NP & SNN & PG & LPCA &PRCUT & CPSCA\\        \hline 
Iris & 0.7302& 0.5503 & 0.7445 & 0.7437 & 0.5601 & 0.7576 & \textbf{0.9216}\\
Flea & 0.6352 & 0.9578 & 0.6531 & 0.6352 & -0.0119 & 0.6352& \textbf{0.9578}\\
Sonar & -0.0001 & 0.0067 & -0.0045 & 0.0066& 0.0014 & -0.0045 & \textbf{0.0733}\\
E-coli & 0.7291 & 0.6662 & 0.4156 & 0.7388 &0.0050 & 0.0399 & \textbf{0.7576}\\
Wine & 0.3204 & 0.8669 & 0.3710 & 0.2926 &0.0008 & 0.1868 & \textbf{0.9167}\\
Soy & 0.5924 & 0.0307 & 0.5924 & 0.5949 &0.0265 & 0.6320 & \textbf{1.0000}\\        
Pima & 0.0023 & 0.0010 & 0.0232 & 0.0308 &0.0023& 0.0153 & \textbf{0.0481}\\
Bupa & -0.0016 & 0.0015 & -0.0037 & -0.0049 & -0.0016 & -0.0016 & \textbf{0.0329}\\
Glass & 0.2104 & 0.2377 & 0.2095 & 0.2024 & 0.2104 & 0.0171 & \textbf{0.2455}\\
BC & 0.8660 & 0.4495 & \textbf{0.8824} & 0.8230 & 0.0026 & 0.8177 & 0.8446\\
Seeds & 0.7166 & 0.6517 & 0.7058 & 0.7031 & 0.4591 & 0.7145 & \textbf{0.7595}\\
Ion & -0.0025 & \textbf{0.6334} & 0.1636 & 0.0723 & 0.0045 & 0.0045 & 0.2399\\

\hline
\end{tabular}             
\end{table}
\begin{table}
\caption{Results of NMI metric on UCI real Data sets}
\label{table:5}
\centering   
\begin{tabular}{|p{1.2cm}|p{1.2cm}|p{1.2cm}|p{1.2cm}|p{1.2cm}|p{1.2cm}|p{1.2cm}|p{1.5cm}|}
\hline
& & & Methods & & & & \\
\hline        
\hline
Data sets & NJW & NP & SNN & PG & LPCA &PRCUT & CPSCA\\        \hline 
Iris & 0.7582&0.6971 & 0.7777 & 0.7661 & 0.8598 & 0.8851 & \textbf{0.9009}\\
Flea & 0.6313 & 0.9470 & 0.6736 & 0.6313 & \textbf{1.0000} & \textbf{1.0000}  & 0.9470\\
BC & 0.7712 & 0.4845 & \textbf{0.8008} & 0.7195 & 0.6287 & 0.7727  & 0.7423\\
Seeds & 0.6949 & 0.6607 & 0.6867 & 0.6812 & \textbf{0.8013} & 0.7425  & 0.7158\\
Soy & 0.7913 & 0.1244 & 0.7913 & 0.7928 & \textbf{1.0000} & \textbf{1.0000}  & \textbf{1.0000}\\   
Pima & 0.0171 & 0.0012 & 0.0078 & 0.0126 & \textbf{0.1091} & 0.1041  & 0.1052\\
Bupa & 0.0136 & 0.0001 & 0.0230 & 0.0055 & \textbf{0.0432} & 0.0324  & 0.0307\\
Glass & 0.3837 & 0.3840 & 0.3665 & 0.3821 & \textbf{0.6270} & 0.6161  & 0.3393\\
Wine & 0.3946 & 0.8482 & 0.4302 & 0.4158 & 0.8821 & \textbf{0.9120}  & 0.8975\\
Ion & 0.0000 & \textbf{0.5893} & 0.1286 & 0.0383 & 0.2387 & 0.2772  & 0.1843\\
Sonar & 0.0004 & 0.0581 & 0.0615 & 0.0098 & \textbf{0.5638} & 0.0449 & 0.096\\
E-coli & 0.6989 & 0.6444 & 0.6166 & 0.6871 & 0.7215 & \textbf{0.7398} & 0.7058\\
\hline
\end{tabular}             
\end{table}
\begin{table}
\caption{Results of CE metric on UCI real Data sets}
\label{table:6}
\centering 
\begin{tabular}{|p{1.2cm}|p{1.2cm}|p{1.2cm}|p{1.2cm}|p{1.2cm}|p{1.2cm}|p{1.2cm}|p{1.5cm}|}
\hline
& & & Methods & & & & \\
\hline        
\hline
Data sets & NJW & NP & SNN & PG & LPCA &PRCUT & CPSCA\\        \hline 
Iris & 0.1067 & 0.3133 & 0.1000 & 0.1000 & 0.0333 & 0.0333 &  \textbf{0.0267}\\
Flea & 0.1622 & \textbf{0.0135} & 0.1622 & 0.1622 & \textbf{0.0135} & \textbf{0.0135} &  \textbf{0.0135}\\
BC & 0.0343 & 0.1645 & \textbf{0.0300} & 0.0458 & 0.0930 & 0.0358 & 0.0401\\
Seeds & 0.1048 & 0.1333 & 0.1095 & 0.1095 & \textbf{0.0571} & 0.0762 &  0.0857\\
Soy & 0.3191 & 0.5957 & 0.3191 & 0.2979 & 0.0213 & 0.0426 &  \textbf{0.0000}\\   
Pima & 0.3477 & 0.3490 & 0.3503 & 0.3451 & 0.3359 & \textbf{0.3060} &  0.3789\\
Bupa & 0.4232 & 0.4290 & 0.4667 & 0.4319 & \textbf{0.3855} & 0.3913 &  0.4000\\
Glass & 0.5093 & 0.5093 & 0.5794 & 0.5000 & 0.3551 & \textbf{0.3271} &  0.5374\\
Wine & 0.4382 & 0.0449 & 0.2921 & 0.3876 & 0.0337 & \textbf{0.0225} &  0.0955\\
Ion & 0.4274 & \textbf{0.0997} & 0.2963 & 0.3590 & 0.3561 & 0.2365 &  0.2422\\
Sonar & 0.4663 & 0.4471 & 0.4952 & 0.4471 & \textbf{0.1010} & 0.3990 &  0.3606\\
E-coli & 0.2321 & 0.2411 & 0.4196 & 0.1994 & 0.1726 & \textbf{0.1548} &  0.2143\\
\hline
\end{tabular}             
\end{table}
\subsection{Handwritten data sets}
The third category of data we have considered is MNIST handwritten digits database given by Lecun et al.\cite{lecun1998gradient}. The database has a training set of 60,000 examples and test set of 10,000 samples. There are 1000 samples for each of the ten digits (0-9). All the samples are images of size 28x28.\\ 
For our experiments, we have considered some of challenging test cases such as \{0,8\}, \{3,5,8\}, \{1,2,3,4\}. We have taken 200 samples of each digit in each test case. The results of comparison with other techniques is presented in Tables \ref{table:7}, \ref{table:8}, \ref{table:9}. In the following subsection we discuss the robustness of our method to the parameter $k$.
\subsection{Parameter sensitivity}
\begin{figure}
\centering
\includegraphics[scale=.5]{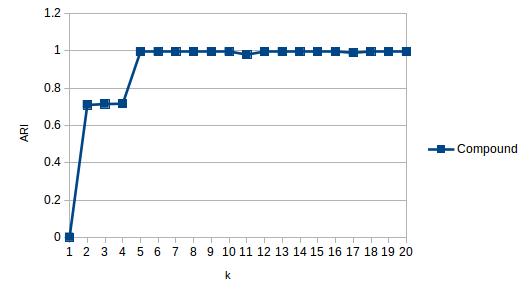}
\includegraphics[scale=.5]{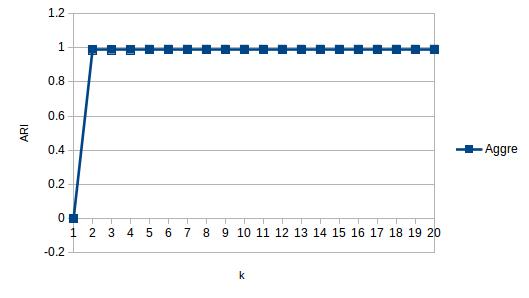}
\includegraphics[scale=.5]{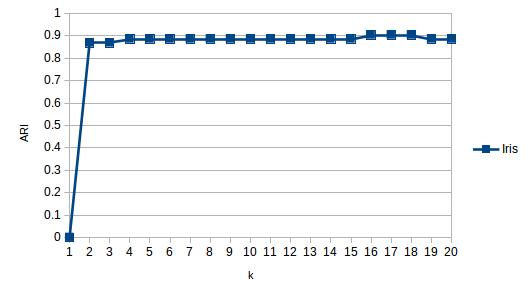}
\includegraphics[scale=.5]{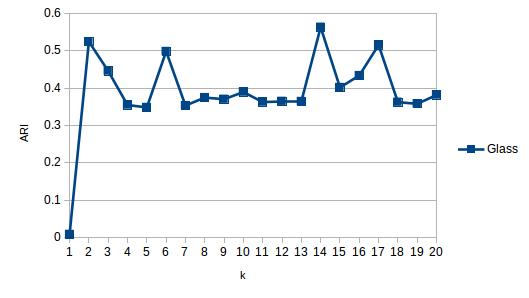}
\caption[Summation Index]{Robustness of CPSC: ARI vs k} 
\label{robust}
\end{figure}

In this section, we have shown how the method is sensitive to the value of k. 
We analyzed, how the clustering accuracy(ARI value) changes with respect to $k$, when $k$ is increased from 1 to 20. Sample plots of ARI versus $k$ are shown in Fig \ref{robust}. 
\begin{table}
\caption{ Results of ARI metric on MNIST}
\label{table:7}
\centering
\begin{tabular}{|p{1.2cm}|p{1.2cm}|p{1.2cm}|p{1.2cm}|p{1.2cm}|p{1.2cm}|p{1.2cm}|p{1.5cm}|}
\hline
& & & Methods & & & & \\
\hline        
\hline
Data sets & NJW & NP & SNN & PG & LPCA &PRCUT & CPSCA\\  
\hline
\{0,8\} & \textbf{1.0000} & 0.9601 & 0.9799 &\textbf{ 1.0000} & 0.9899 & 0.9799  & \textbf{1.0000}\\
\{3,5,8\}  & 0.6190 & 0.5664 & 0.5664 &\textbf{ 0.6498} & 0.5657 & 0.5621 &0.5694\\
\{1,2,3,4\}& 0.3431 & 0.3349 & 0.3349 & 0.4220 & 0.3310 & 0.3512 &\textbf{ 0.4916}\\
\hline
\end{tabular}             
\end{table}
\begin{table}
\caption{Results of NMI metric on MNIST}
\label{table:8}
\centering
\begin{tabular}{|p{1.2cm}|p{1.2cm}|p{1.2cm}|p{1.2cm}|p{1.2cm}|p{1.2cm}|p{1.2cm}|p{1.5cm}|}
\hline
& & & Methods & & & & \\
\hline        
\hline
Data sets & NJW & NP & SNN & PG & LPCA &PRCUT & CPSCA\\  
\hline
\{0,8\} & \textbf{1.0000} & 0.9288 & 0.9594 &\textbf{ 1.0000} &0.9772 &0.9594 & \textbf{1.0000}\\
\{3,5,8\}  & 0.7502 & 0.7367 & 0.7367 & 0.7502 &  0.7502 &0.7418  & \textbf{0.7517}\\
\{1,2,3,4\}& 0.6282 & 0.5327 & 0.5234 & 0.5897 &0.6235 &0.6282  & \textbf{0.6827}\\
\hline
\end{tabular}             
\end{table}
\begin{table}
        \caption{Results of CE metric on MNIST}
        \label{table:9}
        \centering    \begin{tabular}{|p{1.2cm}|p{1.2cm}|p{1.2cm}|p{1.2cm}|p{1.2cm}|p{1.2cm}|p{1.2cm}|p{1.5cm}|}
\hline
& & & Methods & & & & \\
\hline        
\hline
Data sets & NJW & NP & SNN & PG & LPCA &PRCUT & CPSCA\\  
            \hline
            \{0,8\}& \textbf{0.0000} & 0.0100 & 0.0050 & \textbf{0.0000} &0.0025 &0.0050& \textbf{0.0000}\\
            \{3,5,8\} & 0.1633 & 0.3283 & 0.3283 &\textbf{ 0.1433}&0.3367&0.3367  & 0.2967\\
            \{1,2,3,4\} & 0.4313 & 0.4637 & 0.4637 & \textbf{0.3675}&0.5012 &0.4425 & 0.4963\\
            \hline
        \end{tabular}             
    \end{table}
\section{Conclusion}
\label{conclusion}
Traditionally in a spectral clustering algorithm Gaussian weighted distance is used to find the affinity between data points. Using the concepts of conformal prediction such as P-value, we have proposed a novel hybrid similarity metric by incorporating local neighborhood information in construction of pairwise affinity. The results on synthetic, real and handwritten data sets have shown that conformity based similarity between the data points has improved the effectiveness of SC.    
\begin{acknowledgements}
We dedicate our work to Bhagawan Sri Sathya Sai Baba, the founder chancellor of Sri Sathya Sai Institute of Higher Learning, .
\end{acknowledgements}
\bibliographystyle{spmpsci} 
\bibliography{report2}   

\begin{thebibliography}{10}
\providecommand{\url}[1]{{#1}}
\providecommand{\urlprefix}{URL }
\expandafter\ifx\csname urlstyle\endcsname\relax
  \providecommand{\doi}[1]{DOI~\discretionary{}{}{}#1}\else
  \providecommand{\doi}{DOI~\discretionary{}{}{}\begingroup
  \urlstyle{rm}\Url}\fi

\bibitem{arias2017spectral}
Arias-Castro, E., Lerman, G., Zhang, T.: Spectral clustering based on local
  pca.
\newblock The Journal of Machine Learning Research \textbf{18}(1), 253--309
  (2017)

\bibitem{balasubramanian2014conformal}
Balasubramanian, V., Ho, S.S., Vovk, V.: Conformal prediction for reliable
  machine learning: Theory, Adaptations and applications.
\newblock Newnes (2014)

\bibitem{beauchemin2015density}
Beauchemin, M.: A density-based similarity matrix construction for spectral
  clustering.
\newblock Neurocomputing \textbf{151}, 835--844 (2015)

\bibitem{challapower}
Challa, A., Danda, S., Sagar, B.D., Najman, L.: Power spectral clustering.
\newblock hal-01516649v3 pp.~-- (2018).
\newblock \urlprefix\url{https://hal.archives-ouvertes.fr/hal-01516649}

\bibitem{chang2008robust}
Chang, H., Yeung, D.Y.: Robust path-based spectral clustering.
\newblock Pattern Recognition \textbf{41}(1), 191--203 (2008)

\bibitem{cherubin2015conformal}
Cherubin, G., Nouretdinov, I., Gammerman, A., Jordaney, R., Wang, Z., Papini,
  D., Cavallaro, L.: Conformal clustering and its application to botnet
  traffic.
\newblock In: SLDS, pp. 313--322 (2015)

\bibitem{diao2015spectral}
Diao, C., Zhang, A.H., Wang, B.: Spectral clustering with local projection
  distance measurement.
\newblock Mathematical Problems in Engineering \textbf{2015} (2015)

\bibitem{eklund2015application}
Eklund, M., Norinder, U., Boyer, S., Carlsson, L.: The application of conformal
  prediction to the drug discovery process.
\newblock Annals of Mathematics and Artificial Intelligence \textbf{74}(1-2),
  117--132 (2015)

\bibitem{fu2007flame}
Fu, L., Medico, E.: Flame, a novel fuzzy clustering method for the analysis of
  dna microarray data.
\newblock BMC bioinformatics \textbf{8}(1), 3 (2007)

\bibitem{gionis2007clustering}
Gionis, A., Mannila, H., Tsaparas, P.: Clustering aggregation.
\newblock ACM Transactions on Knowledge Discovery from Data (TKDD)
  \textbf{1}(1), 4 (2007)

\bibitem{gu2009improved}
Gu, R., Wang, J.: An improved spectral clustering algorithm based on neighbour
  adaptive scale.
\newblock In: Business Intelligence and Financial Engineering, 2009. BIFE'09.
  International Conference on, pp. 233--236. IEEE (2009)

\bibitem{hubert1985comparing}
Hubert, L., Arabie, P.: Comparing partitions.
\newblock Journal of classification \textbf{2}(1), 193--218 (1985)

\bibitem{jain2005data}
Jain, A.K., Law, M.H.: Data clustering: A user's dilemma.
\newblock PReMI \textbf{3776}, 1--10 (2005)

\bibitem{jordan2004learning}
Jordan, F., Bach, F.: Learning spectral clustering.
\newblock Adv. Neural Inf. Process. Syst \textbf{16}, 305--312 (2004)

\bibitem{kaufman2009finding}
Kaufman, L., Rousseeuw, P.J.: Finding groups in data: an introduction to
  cluster analysis, vol. 344.
\newblock John Wiley \& Sons (2009)

\bibitem{laxhammar2011sequential}
Laxhammar, R., Falkman, G.: Sequential conformal anomaly detection in
  trajectories based on hausdorff distance.
\newblock In: Information Fusion (FUSION), 2011 Proceedings of the 14th
  International Conference on, pp. 1--8. IEEE (2011)

\bibitem{lecun1998gradient}
LeCun, Y., Bottou, L., Bengio, Y., Haffner, P.: Gradient-based learning applied
  to document recognition.
\newblock Proceedings of the IEEE \textbf{86}(11), 2278--2324 (1998).
\newblock \urlprefix\url{http://yann.lecun.com/exdb/mnist/}

\bibitem{li2012constructing}
Li, X.Y., Guo, L.J.: Constructing affinity matrix in spectral clustering based
  on neighbor propagation.
\newblock Neurocomputing \textbf{97}, 125--130 (2012)

\bibitem{Lichman:2013}
Lichman, M.: {UCI} machine learning repository (2013).
\newblock \urlprefix\url{http://archive.ics.uci.edu/ml}

\bibitem{natalianipowered}
Nataliani, Y., Yang, M.S.: Powered gaussian kernel spectral clustering.
\newblock Neural Computing and Applications  (2017).
\newblock \doi{10.1007/s00521-017-3036-2}.
\newblock \urlprefix\url{https://doi.org/10.1007/s00521-017-3036-2}

\bibitem{ng2002spectral}
Ng, A.Y., Jordan, M.I., Weiss, Y.: On spectral clustering: Analysis and an
  algorithm.
\newblock Advances in neural information processing systems \textbf{2},
  849--856 (2002)

\bibitem{rand1971objective}
Rand, W.M.: Objective criteria for the evaluation of clustering methods.
\newblock Journal of the American Statistical association \textbf{66}(336),
  846--850 (1971)

\bibitem{shi2000normalized}
Shi, J., Malik, J.: Normalized cuts and image segmentation.
\newblock Pattern Analysis and Machine Intelligence, IEEE Transactions on
  \textbf{22}(8), 888--905 (2000)

\bibitem{Smith2014Anomaly}
Smith, J., Nouretdinov, I., Craddock, R., Offer, C., Gammerman, A.: Anomaly
  detection of trajectories with kernel density estimation by conformal
  prediction.
\newblock In: IFIP International Conference on Artificial Intelligence
  Applications and Innovations, pp. 271--280. Springer (2014)

\bibitem{strehl2003cluster}
Strehl, A., Ghosh, J.: Cluster ensembles: a knowledge reuse framework for
  combining multiple partitions.
\newblock The Journal of Machine Learning Research \textbf{3}, 583--617 (2003)

\bibitem{verma2003comparison}
Verma, D., Meila, M.: A comparison of spectral clustering algorithms.
\newblock University of Washington Tech Rep UWCSE030501 \textbf{1}, 1--18
  (2003)

\bibitem{vovk2014basic}
Vovk, V.: The basic conformal prediction framework.
\newblock Conformal Prediction for Reliable Machine Learning: Theory,
  Adaptations, and Applications pp. 3--19 (2014)

\bibitem{wong1980asymptotic}
Wong, M.A.: Asymptotic properties of k-means clustering algorithm as a density
  estimation procedure  (1980)

\bibitem{xu2005survey}
Xu, R., Wunsch, D.C.: Survey of clustering algorithms  (2005)

\bibitem{yang2011spectral}
Yang, P., Zhu, Q., Huang, B.: Spectral clustering with density sensitive
  similarity function.
\newblock Knowledge-Based Systems \textbf{24}(5), 621--628 (2011)

\bibitem{ye2016robust}
Ye, X., Sakurai, T.: Robust similarity measure for spectral clustering based on
  shared neighbors.
\newblock ETRI Journal \textbf{38}(3), 540--550 (2016)

\bibitem{zahn1971graph}
Zahn, C.T.: Graph-theoretical methods for detecting and describing gestalt
  clusters.
\newblock IEEE Transactions on computers \textbf{100}(1), 68--86 (1971)

\bibitem{zelnik2004self}
Zelnik~Manor, L., Perona, P.: Self tuning spectral clustering.
\newblock In: Advances in neural information processing systems, pp. 1601--1608
  (2004).
\newblock
  \urlprefix\url{http://www.vision.caltech.edu/lihi/Demos/SelfTuningClustering.html}

\bibitem{zhang2011local}
Zhang, X., Li, J., Yu, H.: Local density adaptive similarity measurement for
  spectral clustering.
\newblock Pattern Recognition Letters \textbf{32}(2), 352--358 (2011)

\end{thebibliography}
\section*{Appendix}
In this section various metrics which were used in the work, namely: ARI, NMI, CE will be defined. 
\subsection*{Adjusted Rand Index(ARI)}
\cite{hubert1985comparing} defines ARI using the Contingency table defined as:\\
Given a set $D$ of p elements, and two groupings or partitions (e.g. clusterings) of these points, namely $ I=\{I_{1},I_{2},\ldots ,I_{r}\}$ and $ J=\{J_{1},J_{2},\ldots ,J_{s}\}$, the overlap between $ I$ and $ J$ can be summarized in a contingency table $ \left[pt_{ij}\right]$ where each entry $ pt_{ij}$ denotes the number of objects in common between $ I_{i}$ and $ J_{j}$: 
$ pt_{ij}=|I_{i}\cap J_{j}|$
    
\begin{center}
\begin{tabular}{ |c|c c c c|c| } 
\hline
$I \backslash J$   & $J_{1}$ & $J_{2}$ & $\ldots$ & $J_{s}$  & $Sums$  \\ \hline
$ I_{1} $ & $ pt_{11} $ & $ pt_{12} $ & $  \ldots $ &$  pt_{1s} $ & $ t_{1} $\\
$ I_{2} $ & $ pt_{21} $& $ pt_{22} $ &$   \ldots  $& $ pt_{2s} $ & $ t_{2} $\\
$\vdots$  & $\vdots$  & $\vdots$  & $\ddots$  & $\vdots$  & $\vdots$  \\
$I_{r}$ & $pt_{r1}$ & $pt_{r2}$ &   $\ldots$ & $pt_{rs}$ &  $t_r$\\ \hline
$Sums$ & $l_{1}$ & $ l_{2}$ & $\ldots$ & $l_{s}$ & \\
\hline
\end{tabular}
\end{center}
The adjusted form of the Rand Index, the Adjusted Rand Index, is \\
\begin{equation}
{\text{AdjustedIndex}}={\frac {{\text{Index}}-{\text{ExpectedIndex}}}{{\text{MaxIndex}}-{\text{ExpectedIndex}}}} 
\end{equation}
, more specifically
\begin{equation}
ARI={\frac {\sum _{ij}{\binom {pt_{ij}}{2}}-[\sum _{i}{\binom {t_{i}}{2}}\sum _{j}{\binom {l_{j}}{2}}]/{\binom {p}{2}}}{{\frac {1}{2}}[\sum _{i}{\binom {t_{i}}{2}}+\sum _{j}{\binom {l_{j}}{2}}]-[\sum _{i}{\binom {t_{i}}{2}}\sum _{j}{\binom {l_{j}}{2}}]/{\binom {p}{2}}}}
\end{equation}
where $ pt_{ij},t_{i},l_{j}$ are values from the contingency table.
\subsection*{Normalized Mutual Information(NMI)}
Normalized Mutual Information(NMI) is defined as in \cite{strehl2003cluster}:\\
\begin{equation}
NMI=\frac{    \sum_{k=1}^{m}\sum_{l=1}^{q}\big( num_{k,l}\times log\big(\frac{n\times num_{k,l}}{num_k\times  \hat{num}_l}\big)\big)}{\sqrt{\big(\sum_{k=1}^{m}num_k\times  log(\frac{num_k}{n})\big)\times \big(\sum_{l=1}^{q}\hat{num}_l\times  log(\frac{\hat{num}_l}{n})\big)}}
\end{equation}
where n is the total number of points in data,  $num_k$ denotes the number of datapoints contained in cluster $C_k(1\leq k \leq m)$, $\hat{n}_i$ is the number of data belonging to the $l^{th}$ class $(1\leq l \leq q)$ and $n_{i,j}$ denotes the number of data that are in the intersection between $C_i$ and the $j^{th}$ class. 
\subsection*{Clustering error:}
If $Conf$ is defined as the confusion matrix of two clusterings.
\begin{equation}
Conf(n_{true},n)=|D^{true}_{n_{true}}\cap D_n|
\end{equation}
Confusion($Conf$) is the number of common points between clustering produced($ n $) and true clustering($n_{true}$). According to \cite{verma2003comparison}, clustering error(CE) is defined as :
\begin{equation}
CE(D,D^{true})=\Big(\sum_{n_{true}}\sum_{n\neq n_{true}} Conf(n_{true},n)\Big)/n
\end{equation}
where n is the total number of points.\\
Due to the possibility of renumbering in the output clustering, cluster 2 could be allocated cluster 1 by the algorithm. Hence CE is considered as a minimum value of all possible combinations of numberings. This method has large number of computations. A maximum weighted bipartite matching problem based modeling is applied and its solution is found using linear programming, in order to reduce computation.    
\end{document}